\documentclass[graybox]{svmult}

\usepackage{mathptmx}       
\usepackage{helvet}         
\usepackage{courier}        
\usepackage{type1cm}        
\usepackage{makeidx}         
\usepackage{graphicx}        
\usepackage{multicol}        
\usepackage[bottom]{footmisc}
\usepackage{comment}
\usepackage{url}
\usepackage{microtype}       

\makeindex             

\usepackage{babel}
\usepackage{color}
\usepackage{xcolor}
\usepackage{graphicx}
\usepackage{mathtools}

\graphicspath{ {./imgs/} }

\newcommand{\jmf}[1]{} 
\newcommand{\bdk}[1]{}
\newcommand{\jzc}[1]{}
\newcommand{\jic}[1]{}
\newcommand{\mep}[1]{}
\newcommand{\shd}[1]{}

\newcommand{\jmflater}[1]{} 

\begin{document}

\title*{Dependency Dialogue Acts — Annotation Scheme and Case Study}
\author{Jon Z. Cai, Brendan King, Margaret Perkoff, Shiran Dudy, Jie Cao, Marie Grace, Natalia Wojarnik, Ananya Ganesh, James H. Martin, Martha Palmer, Marilyn Walker and Jeffrey Flanigan}
\institute{Jon Z. Cai \at University of Colorado Boulder, 1134 CO-93, Boulder, CO 80302, \email{jon.z.cai@colorado.edu}
\and Brendan King,  Jeffrey Flanigan \at University of California Santa Cruz, 1156 High St., Santa Cruz, CA 95064, \email{bking2@ucsc.edu,jmflanig@ucsc.edu}}
%
%

\maketitle

\abstract{In this paper, we introduce Dependency Dialogue Acts (DDA), a novel framework for capturing the structure of speaker-intentions in multi-party dialogues. DDA combines and adapts features from existing dialogue annotation frameworks, and emphasizes the multi-relational response structure of dialogues in addition to the dialogue acts and rhetorical relations. It represents the functional, discourse, and response structure in multi-party multi-threaded conversations. A few key features distinguish DDA from existing dialogue annotation frameworks such as SWBD-DAMSL and the ISO 24617-2 standard. First, DDA prioritizes the relational structure of the dialogue units and the dialog context, annotating both dialog acts and rhetorical relations as response relations to particular utterances. Second, DDA embraces overloading in dialogues, encouraging annotators to specify multiple response relations and dialog acts for each dialog unit. Lastly, DDA places an emphasis on adequately capturing how a speaker is using the full dialog context to plan and organize their speech.  With these features, DDA is highly expressive and recall-oriented with regard to conversation dynamics between multiple speakers.  In what follows, we present the DDA annotation framework and case studies annotating DDA structures in multi-party, multi-threaded conversations.}

\section{Introduction}
\label{sec:1}
Discourse analysis has become an increasingly popular problem in natural language processing.  Broadly, discourse analysis for dialog involves observing a conversation between two or more individuals and understanding the information that is being exchanged, both explicitly and implicitly.  One of the goals of dialogue analysis systems is to be able to understand the intents of the parties involved. This problem becomes more difficult when analyzing conversations between more than two people in an open environment. In these settings, side conversations and non-discourse events can interrupt an ongoing conversation. These multi-party multi-threaded scenarios are ones that we encounter on a daily basis. Additionally, these complex conversations contain abundant information that indicates the relationship between the speakers, their moods, their likes and dislikes, as well as their intentions.  Presently, we are seeing an influx of conversational agents that are attempting to mimic our ability to not only interpret this information correctly but to generate an appropriate response to it as well.

Previous research in discourse analysis has led to a variety of annotation schemes that attempt to capture different aspects of conversations. One of the foundational schemes in this space is the Switchboard DAMSL~\cite{jurafsky1997switchboard}
which annotates conversations at the utterance level based on their corresponding dialog acts.  Dialog acts are used to represent the intention of the speaker, such as asking a \emph{Question} or expressing a \emph{Statement-Opinion}.  Other alternative schemas include Rhetorical Structure Theory (RST) and shallow discourse relation frameworks such as Penn Discourse Tree Bank~(PDTB) which are frequently used to analyze text structure and coherence.
These schemes have proven extremely valuable in analyzing dialogue but can encounter unique challenges in multi-party, open-environment settings such as our domain of interest -- classroom conversations, where conversation threads interweave and are interrupted by events outside the discourse.

With this in mind, we set out to design a discourse analysis scheme that is able to track the intentions of multiple speakers while preserving the relational information from one turn of the dialogue to the next.  Furthermore, we want our scheme to be sufficiently useful for a conversational agent to generate an appropriate response in a multiparty conversation, which aligns with our goal of creating more explainable and controllable dialogue generation agents. Prior work has shown rhetorical structures and dialog acts can improve controllability and explainability in response generation in dialog agents~\cite{reed-etal-2018-neural, balakrishnan-etal-2019-constrained, li-etal-2021-self}.

Our proposed Dependency Dialogue Act (DDA) annotation scheme builds upon previous work on discourse annotation by merging different features from existing schemes into a single system that captures a large amount of conversational context while minimizing annotator effort.  One of the primary goals is the ability to preserve both rhetorical and response relations between different turns in the utterance.

Additionally, we want to embrace the inherent overloading nature of conversations by enabling annotators to select multiple labels per utterance where appropriate.  Finally, the DDA scheme anchors the speaker's intention with context.

The goal of this paper is to define the Dependency Dialogue Act (DDA) annotation scheme for discourse analysis and investigate its effectiveness in the context of multi-threaded multi-party conversations. In Section~\ref{sec:formal_dda}, we define the response structure of DDA and present the tagset composed of two class types: Dialog Acts and Rhetorical Relations. We demonstrate the usefulness of the DDA scheme with examples from diverse conversation settings throughout. We discuss applications in the dialog analysis space in Section~\ref{sec:applications}. We briefly review prior work on discourse annotation schemes and highlight key features that each of them captures in Section~\ref{sec:related-work}. The ability to adequately interpret multi-party multi-threaded conversations has significant implications for conversational technology across many domains; we hope that the DDA scheme is a step towards capturing more of the critical information present in these settings.

\begin{figure}[b!]
\centering
\includegraphics[width=\columnwidth]{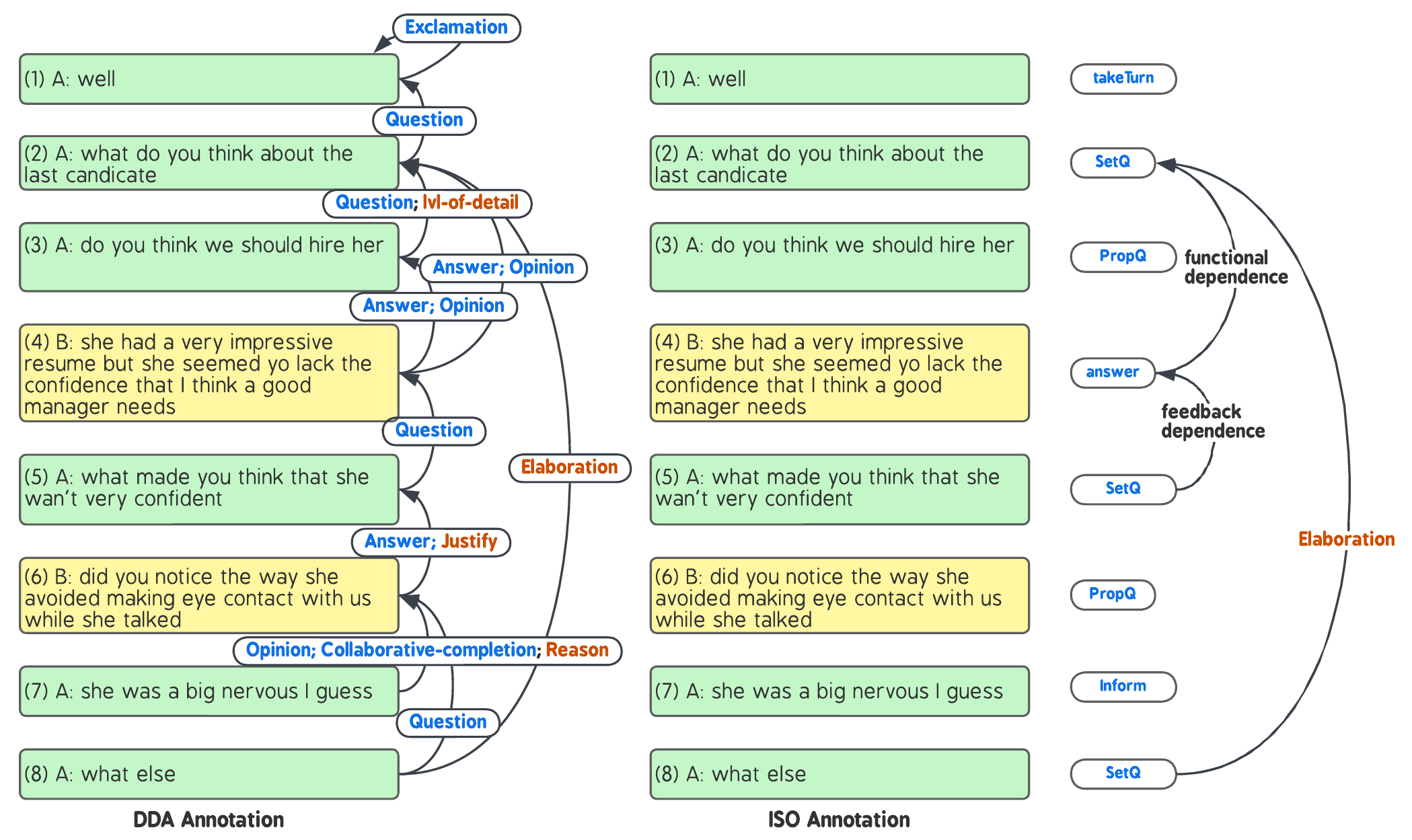}
\caption{An example from the DialogBank corpus as originally annotated with ISO 24617-2 (right) and with Dependency Dialog Act (DDA) (left). Dialog act labels are in blue, and rhetorical relations are in red. The ISO annotation contains a functional dependency ($4 \to 2$), a feedback dependency ($5 \to 4$) and a rhetorical relation ($8 \to 2$), giving context for units $4,5, \& 8$ respectively. DDA annotations are context-oriented, explicitly marking context with response dependencies for all units in a dialogue. This broader view of dialogue structure leads to a fully connected dialog thread that can be disentangled from others in the multi-party setting, by design. DDA annotations are also recall-oriented, encouraging the use of multiple labels for multi-function conversation units.}
\label{fig:dda-iso}
\end{figure}

\section{Motivation}
\label{sec: motivation}

\begin{figure*}[t!]
    \centering
    \label{fig:depcompare}
    \includegraphics[width=\textwidth]{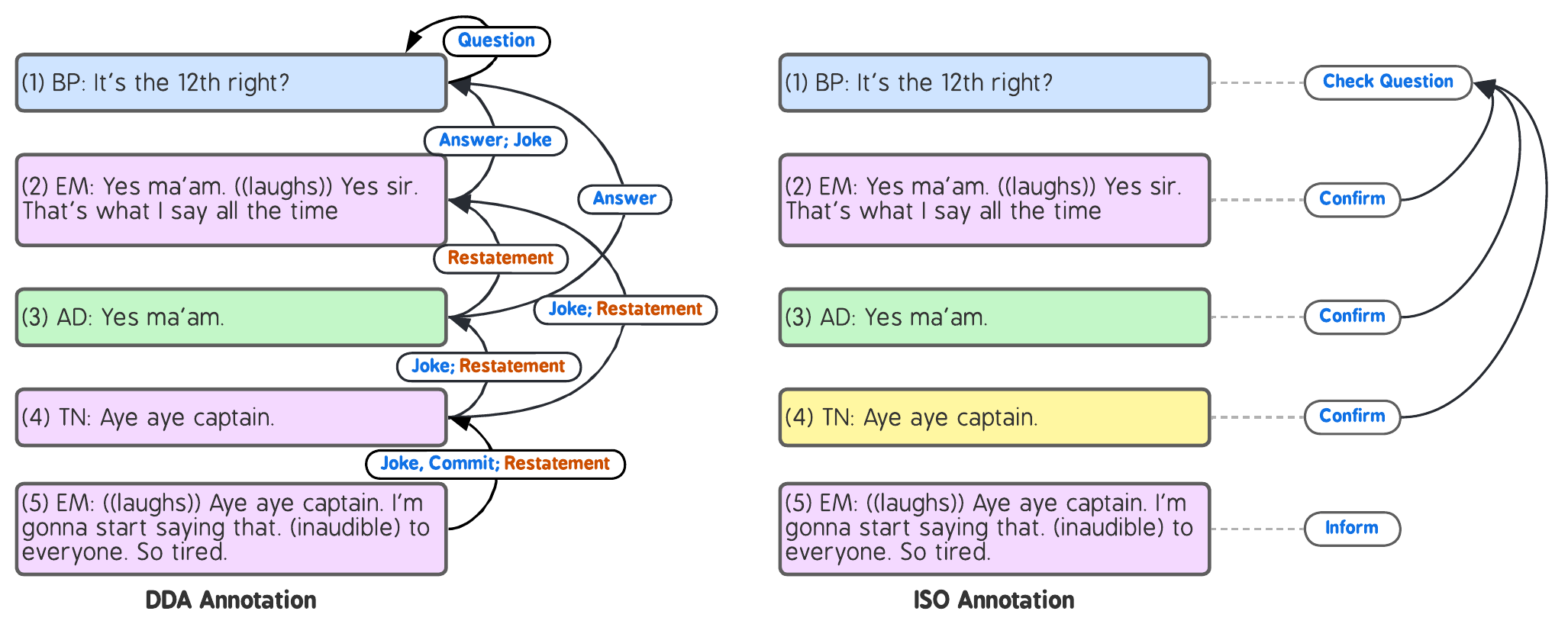}    
    \caption{An example from our classroom setting, annotated with DDA (left) and ISO (right). DDA annotation embraces overloading of multi-purpose units such as in $2,...5$, yielding a comprehensive, recall-oriented structure.$^1$}
\end{figure*}

\begin{figure}[h!]
    \centering
    \includegraphics[width=\columnwidth]{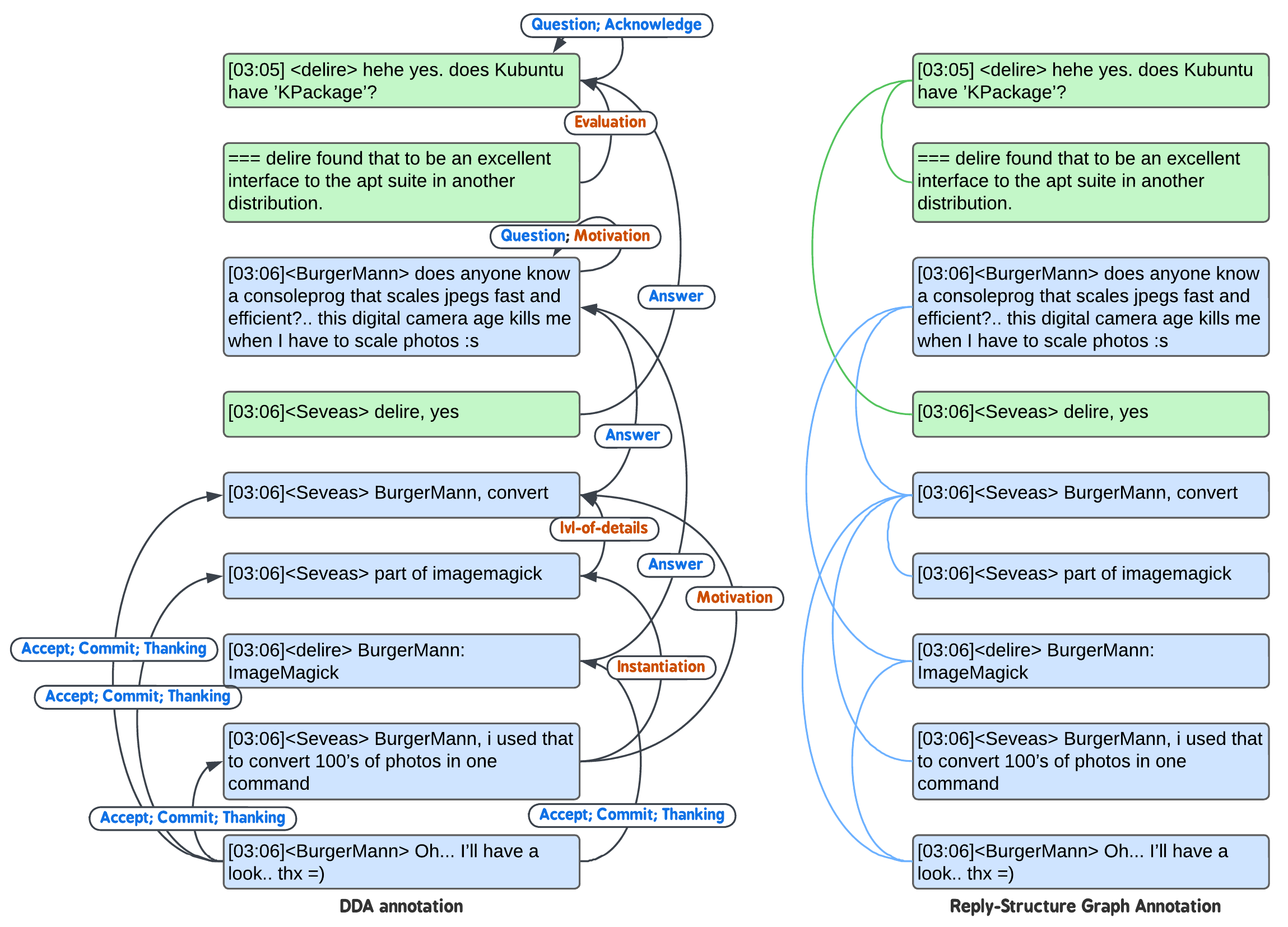}
    \caption{Annotation of DDA (left) as compared to a reply-structure graph (right) from the Ubuntu-IRC corpus \cite{kummerfeld_large-scale_2019}. DDA contains all the edges in Ubuntu-IRC's reply structure, with labeled dialog acts and rhetorical relations. DDA can contain additional edges for rhetorical relations and dialog acts. Both structures yield separable conversation threads as connected components.\label{fig:ubuntu-irc-example}}
    
\end{figure}

We motivate our Dependency Dialog Act~(DDA) annotation scheme with three examples, shown in Figs.~\ref{fig:dda-iso}-\ref{fig:ubuntu-irc-example}. DDA aims to capture as much information about the interrelationships between utterances as possible while also representing the multiple dialog acts and rhetorical relations that a single utterance can have.  DDA captures the response structure, dialog acts, and rhetorical relations in one integrated graph structure.  Compared to ISO 24617-2, DDA has more dialog acts for each utterance, and more relations (Figs. \ref{fig:dda-iso} and \ref{fig:depcompare}).\footnote{In Fig.~\ref{fig:depcompare}, (1) poses a question. (2) can be considered an ``Answer'' to (1). Similarly, the units that follow restate this joke while answering the question posed in (1), each having a different functional response dependency to (1) and rhetorical response dependency on (2), (3), and/or (4). Despite the appearance of an answer to (1), the intent of the speaker in (5) is more likely to be participation in the joke, as the question has already been answered by the same speaker.}
DDA's edges represent response relations, with conversation threads forming connected components, similar to reply-structure graphs in the Ubuntu-IRC corpus (Fig. \ref{fig:ubuntu-irc-example})~\cite{kummerfeld_large-scale_2019}.

\section{Dependency Dialog Act}
\label{sec:formal_dda}
We propose the Dependency Dialog Act (DDA) annotation scheme to capture a broad range of speaker intentions and their relationships to the dialogue context in the multi-party setting. We emphasize the following key design philosophies:
\begin{enumerate}
    \item DDA is context oriented and recommends annotators think from a relational perspective. This is reflected in DDA by annotating dialog acts and relations on response edges to the surrounding context, rather than on dialog units (see end of Section~\ref{sec:formalizing_resp_dep_structures}).
    \item DDA is recall-oriented and encourages annotators to put all response relations that fit for the given context. It embraces overloading as an important feature of the framework (see end of Section~\ref{sec:overloading}).
    \item DDA pays attention to speaker intentions, trying to capture both the purpose of speech and ``how'' a speaker plans and arranges their speech conditioned on the context.  This philosophy is reflected in the design decisions of DDA.\jmflater{perhaps we should elaborate somewhere else in the paper}
\end{enumerate}

DDA aims at capturing \textit{speaker intention} as a key feature. ``Intention'' is a widely studied concept in philosophy~\cite{Anscombe1979-ANSIAI-2, vermazen1985essays}, theory of action, and logic~\cite{Jeffrey1965-JEFTLO-2}. We follow the functionalist philosophy~\cite{cohen1990intentions}, defining \textbf{intentions} as operational plans either in our mind or can be entailed by current actions. 
For example, when a speaker provides a ``action-directive'' utterance, the speaker's entailed plan is to have certain actions performed. Then, when providing further ``elaboration,'' the speaker's plan is to make the existing statement more convincing or clear. DDA uses an enhanced dialog acts set from SWBD-DAMSL as the basis to describe actions performed, and enhanced discourse relations as the basis to describe discourse purpose plans. It embeds intentional information and context into the labeled dependency edges.

We introduce the Dependency Dialog Act annotation scheme in two sections:
In Section \ref{sec:formalizing_resp_dep_structures}, we define our response dependency relations between units of dialog (see Slash Units, below).
In Section \ref{sec:DADR_classes}, we describe the adaptation of existing tag schema for dialogue acts and rhetorical relations to form the basis of DDA's intention space.\\

\noindent\textbf{Dialog Units of Annotation: Slash Units}\label{ssec: slash-unit} - Similar to \textit{functional segments} in ISO 24617-2 standard and \textit{elementary discourse units} (EDUs) in RST, we assume that a dialogue is broken up into units for annotation. Following the SWBD-DAMSL annotation scheme, we term these \textbf{slash units}.

\subsection{DDA Edges: Response Dependencies}  

\label{sec:formalizing_resp_dep_structures}

The edges in DDA indicate response relationships between slash units. Specifically, for a slash unit of interest, a \textbf{response dependency} is a directed edge from the unit of interest to the slash unit it depends on or originated from conversationally. When a slash unit $u_i$ has no unit to relate to in the prior context, we use a \textbf{self-pointing dependency} $u_i \to u_i$ to specify the start of a new thread of conversation.

DDA takes an expansive view of response relations between slash units which encompasses the functional, rhetorical, and reply relationships in other frameworks:
\begin{itemize}
\item \textbf{Functional dependency}:\footnote{In addition to functional dependence, the ISO 24617-2 standard defines feedback dependencies for particular feedback acts. Feedback acts largely correspond to backward-communicative-function dialogue acts in SWBD-DAMSL, which we adapt for use in DDA. Thus, we consider feedback dependencies as similar to functional ones, where the interpretation of the slash-unit and label heavily rely on the dependent unit(s).} the meaning of a dialogue act for a local slash unit depends crucially on a particular slash-unit in the dialogue context, such as how an Answer depends on a Question \cite{bunt-etal-2012-iso}.
\item \textbf{Rhetorical relations}: the coherent organization of two slash units, for example labeling units which elaborate on or contrast with previous units \cite{mann1988rhetorical}. Also known as discourse relations.
\item \textbf{Response} or \textbf{continuation dependency} represents continuation of a conversation thread but no explicit functional or rhetorical dependencies between two slash units.
\end{itemize}

In DDA, conversation threads form separate connected components in the DDA annotation graph.  Consider Fig. \ref{fig:dda-iso}, which includes DDA annotations for a snippet from the DialogBank corpus as originally annotated with the ISO 24617-2 standard \cite{bunt_dialogbank_2016}. 
While the ISO annotation includes multiple relation types, some dialogue units in the conversation thread remain disconnected from the structure. While two-party dialogues like this one often follow a single linear thread, this is not always the case\cite{du_discovering_2016}. For example, coherent threads can overlap and might require disentangling for further analysis, as seen in Figs. \ref{fig:ubuntu-irc-example}, \ref{fig:dda-iso-classroom} and \ref{fig:catan-offi}. \jmf{should we talk about how ISO cannot have a functional or feedback relation for some kinds of dialog acts, such as questions, which is why the example has fewer edges?}\jzc{maybe ``this is mainly due to the ``thinking functionally'' nature of ISO''?}

We annotate dialog acts on response edges rather than on slash units. This is in contrast with most previous annotation schemes for dialog acts such as SWBD-DAMSL and ISO.  The benefit of our approach is that it explicitly labels the context for each dialog act. 
For example, in Fig. \ref{fig:dda-iso-classroom}, utterance 32 contains a question, asking ``who wants to go first?". In the DDA annotation, the context is explicitly marked by the response dependency, such that going first can be understood as a leading discussion of the first question in their packet. In the ISO annotation, this context would need to be inferred from the dialogue history, which may be difficult as many of the nearest slash units belong to a different conversation thread.

\begin{table}[t]
\newlength{\width}
\width.155\linewidth

\centering
\begin{tabular}{p{\width} p{.12\linewidth} p{.12\linewidth} p{\width}p{\width}p{\width}}
                                 & DDA & ISO & Ubuntu-IRC & STAC & SWBD-DAMSL \\ \hline
Dialog acts                       & yes & yes  & no & yes (limited) &  yes \\ \hline
Discourse relations               & PDTB+ & subset of~PDTB  & no & task specific &  no \\ \hline
Reply structure              & yes & partial  & yes & no & no \\ \hline
Functional dependence & yes & yes  & no & no & partial \\ \hline
\end{tabular}
\caption{Annotation Scheme Features Comparison. \textbf{Dialog act} and \textbf{PDTB} rows represent whether a scheme uses this set. \textbf{reply structure, continuation and functional dependence} denote three types of dependence structure defined in Section \ref{sec:formalizing_resp_dep_structures}; \textit{partial} means only a subset of the feature can be annotated under a scheme. STAC refers to the annotation scheme~\cite{asher-etal-2016-discourse}.\label{table: framework-comparison}} 
\end{table}

\subsection{DDA Tagset: Dialog Act and Discourse Relation Classes}
\label{sec:DADR_classes}

Many dialogue annotation frameworks label conversation units from one of two perspectives. First, there are frameworks for labeling the function or ``act'' of a dialog unit of interest, including DAMSL, SWBD-DAMSL, and the ISO 24617-2 standard. Other frameworks aim to model the discourse relations between units, drawing from Rhetorical Structure Theory~\cite{Mann1988RhetoricalST, stent-2000-rhetorical} or Segmented Discourse Representation Theory~(SDRT)~\cite{asher2003logics}. Since we want to capture speaker intentions, we aim to capture both categories of these phenomena in multi-party dialogue in a single annotation scheme, by adapting dialog acts from the SWBD-DAMSL scheme \cite{jurafsky1997switchboard} and discourse relations from the Penn Discourse Tree Bank 3.0 scheme \cite{AB2/SUU9CB_2019}. 

Though relatively few schemes attempt to unify these approaches, ours is not the first. 
In particular, the ISO 24617-2 standard includes dialog acts as well as an additional dimension for rhetorical relations, most commonly annotated with the DR-CORE\footnote{While corpora annotated with ISO typically use the DR-CORE rhetorical relations, the guideline itself does not actually specify this, and any set which relates dialog unit pairs may be used.} relation set 
\cite{bunt-etal-2012-iso, Bunt2016ISOD}. While one could annotate multi-party dialogue with the ISO standard by using the finer-grain PDTB relations in the rhetorical dimension, we found this to not fit our distinct approach to the structures described previously, which departs significantly from the ISO annotation guidelines.\jmflater{perhaps move to intro?}\\


\noindent\textbf{Dialogue Act Set:} DDA's dialog act set covers 40 out of the 42 most frequently used Dialog Act (DA) classes from the SWBD-DAMSL scheme. 26 out of the 40 classes are kept with the original definition and class name, while the remaining 15 are collapsed into coarser classes. This leads us to 31 DA classes. The most noticeable merger of SWBD-DAMSL DA classes is from the ``question'' and ``answer'' DAs. We replaced 5 classes of ``answer'' type from SWBD-DAMSL with a single ``answer'' tag and 8 ``question'' DAs from SWDB-DAMSL with 3 coarser ``question'' classes. This is because most of the sub-type ``question'' and ``answer'' tags can be resolved from the lexical level analysis. Additionally, we add ``joke'' as a new DA to cover the social acts in our domain of interest. In this regard, the taxonomy of DDA's dialogue acts labels still follow SWBD-DAMSL's hierarchy with 6 top-level categories, bolded below.
We list DDA's dialogue acts set with this hierarchy as follows:
\begin{itemize}
\item \textbf{Statements}: Statement, Opinion\jmf{, Sarcasm? Lyn says sarcasm is a rabbit-hole}
\item \textbf{Communicative Status}: Self-talk, Abandoned
\item \textbf{Backward-Communicative Functions}: Answer, Stalling, Accept, Reject, \\ Collaborative Completion, Appreciation, Downplayer, Sympathy, Acknowledge, signal-non-understanding
\item \textbf{Forward-Communicative Function}: Task-Management, Offer, \\ Action-Directive, Commit, Question/Info-request, Open-Question, Rhetorical-Question, Apology, Thanking, Exclamation, Explicit-performative, Welcome
\item \textbf{Information Level}: Greeting, Correction, Conventional-closing
\item \textbf{Other}: Hedge, Joke\jmf{, Put-down?}
\end{itemize}


\noindent\textbf{Rhetorical Relation Set:} DDA uses discourse relations from PDTB expanded with some extra relations. 
Aside from dialog acts, discourse relations are very useful for describing speaker intentions, especially speech organizational intentions that dialog acts do not cover. We use the discourse relations set from PDTB 3.0, but extended it with some finer-grained relations for the ``Contingency'' and ``Expansion'' types. For the ``Contingency'' class, we add 4 more asymmetric sub-types for Cause (``Justify'', ``Motivation'', ``Enablement'' and ``Evaluation''). Similarly, we extended the ``Expansion'' class with 3 more relations (``Process-step'', ``Object-attribute'' and ``List''), which are inspired by Amanda Stent's work on RST in Dialog~\cite{stent-2000-rhetorical}. We added these relations because we found them to be useful distinctions in our conversational datasets, in which students discuss, collaborate and negotiate with each other. DDA still leverages the benefits of PDTB's taxonomy hierarchy with this extension.

For comparison, we list the discourse relations adapted in DDA as well as discourse relations from other common frameworks in Table~\ref{table: discourse-relations}.

\begin{table}[t]
\newlength{\cwidth}
\cwidth.19\linewidth

\centering
\begin{tabular}{p{\cwidth} p{\cwidth} p{\cwidth}p{\cwidth}p{\cwidth}}
Category & ISO-DR-core  & PDTB3.0 & SDRT & DDA \\
            \hline
Temporal    & Async, Sync & Async, Sync, Precedence, Succession & Narration, Precondition & Async, Sync, Before, After \\
\hline
Contingency & Cause, Condition, Neg-Condition, Purpose & Cause, Cause+Belief, Cause+SA, Condition, Neg-Condition, Purpose, Reason, Result      & Explanation, Result, Consequence & Cause, Justify, Motivation, Condition, Neg-Condition, Purpose, Enablement, Reason, Result, Evaluation \\
\hline
Comparison  & Contrast, Similarity, Concession & Contrast, Similarity, Concession, Concession+SA & Consequence, Explanation, Contrast, Parallel    & Contrast, Similarity, Concession \\
\hline
Expansion   & Exception, Conjunction, Disjunction, Substitution, Manner, Elaboration, Restatement, Expansion, Exemplification 
            & Instantiation, Level-of-details, Substitution, Equivalence, Disjunction, Exception, Conjunction, Manner 
            & Continuation, Alternation, Elaboration, Background, Commentary Attribution, Source 
            & Expansion, Instantiation, Level-of-details, Substitution, Restatement, Summary, Disjunction, Exception, Conjunction, Manner, Process-step, Object-attribute 
\end{tabular}
\caption{A comparison of discourse relations across frameworks. We choose to replace some of PDTB's relation name from PDTB2.0 for ease of memory such as ``Precedence''=``Before'', ``Succession''=``After'', ``equivalence''=``restatement'', assigning them identical definition.}
\label{table: discourse-relations}
\end{table}

DDA edges always point backward in the conversation (from a slash unit to another slash unit that it is responding to) or are self-edges.
In order to support this directionality without losing expressive power, we make use of the dual discourse relations introduced in PDTB 3.0, such that any asymmetric relationship is annotated from the context of a reply without changing the meaning of the response dependency structure. For instance, if a unit $A$ is a ``Reason'' for a future unit $B$, this $A \rightarrow B$ can be equivalently annotated as $A \leftarrow B$ such that $B$ is a ``Result'' of $A$.
As for asymmetric relations that can be verbified, DDA uses the active or passive voice of the verb to encode the directionality. For example, $C\xrightarrow[]{\text{Enabling}}D$ is equivalent to $C\xleftarrow[]{\text{Enabled}}D$ . This can be read naturally with English: $C\xrightarrow[]{\text{Enabling}}D$ is read ``C is enabling D'' and $C\xleftarrow[]{\text{Enabled}}D$ is read ``D is enabled by C''.\\

\noindent\textbf{Overloading and Multi-functionality:}
\label{sec:overloading}
As part of DDA’s recall-oriented annotation philosophy, we embrace multi-edges to encode the overloading of responsiveness. For example, in Fig. 2, utterance (2) can be considered a “reply” given the only context utterance in this example is a question from utterance (1), such that the reply-to edge is also a response dependency edge. As the conversation proceeds, the intention rendered in utterance (4) shifted away from being an “Answer” to utterance (1) (the marginal information gain of yet another same answer to the previous question diminishes) and therefore serves more toward rhetorical functions instead of communicative ones.

\section{Applications}
\label{sec:applications}

\begin{figure}[t]
\centering
\includegraphics[width=\columnwidth]{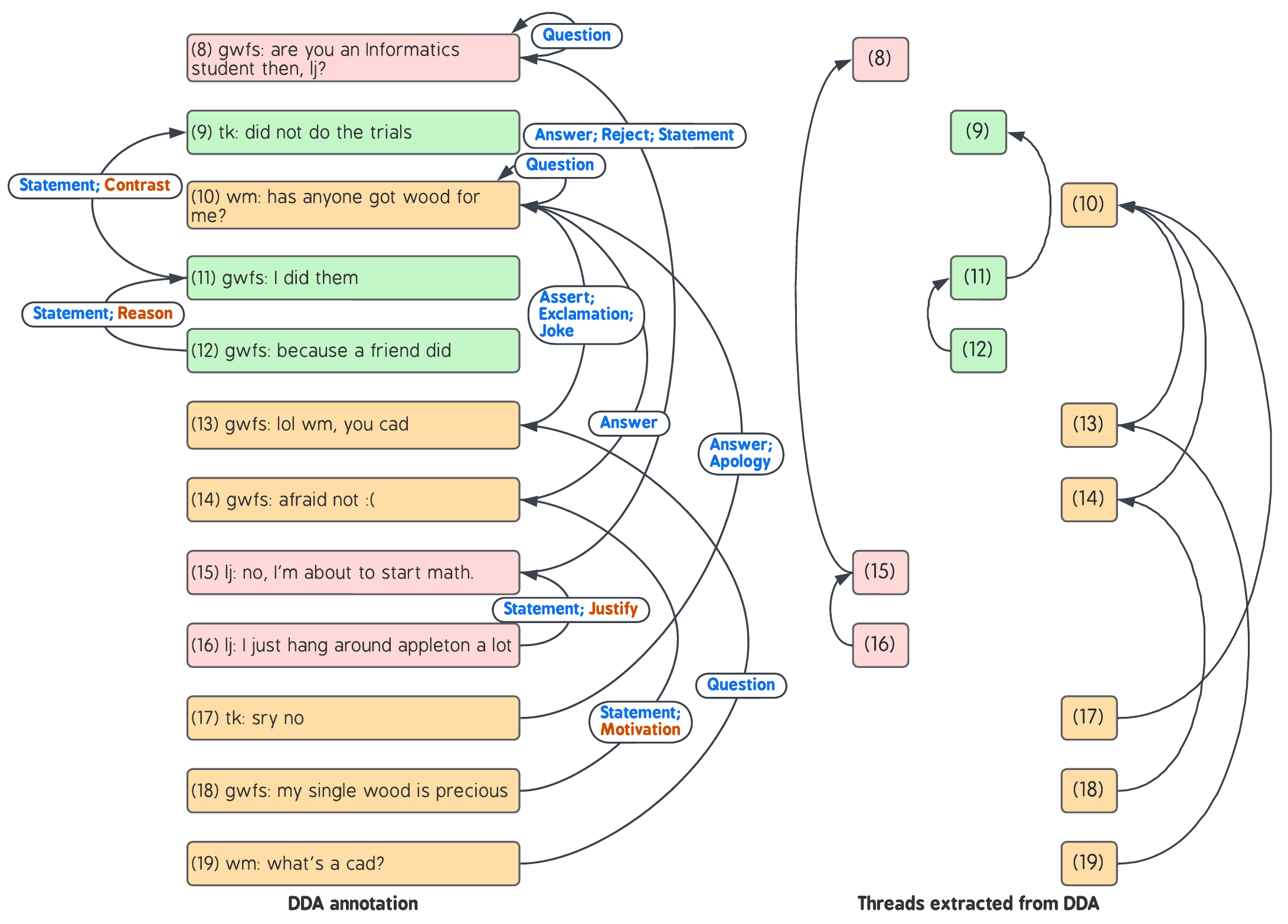}
\caption{DDA annotation on an example in the STAC corpus~\cite{asher-etal-2016-discourse} showing that DDA connects conversation threads into connected components; On the left, we have DDA annotation for the conversation snippet; On the right, we highlight the extracted threads. \label{fig:thread_example}}
\end{figure}
\begin{figure}[t]
\centering
\includegraphics[width=\columnwidth]{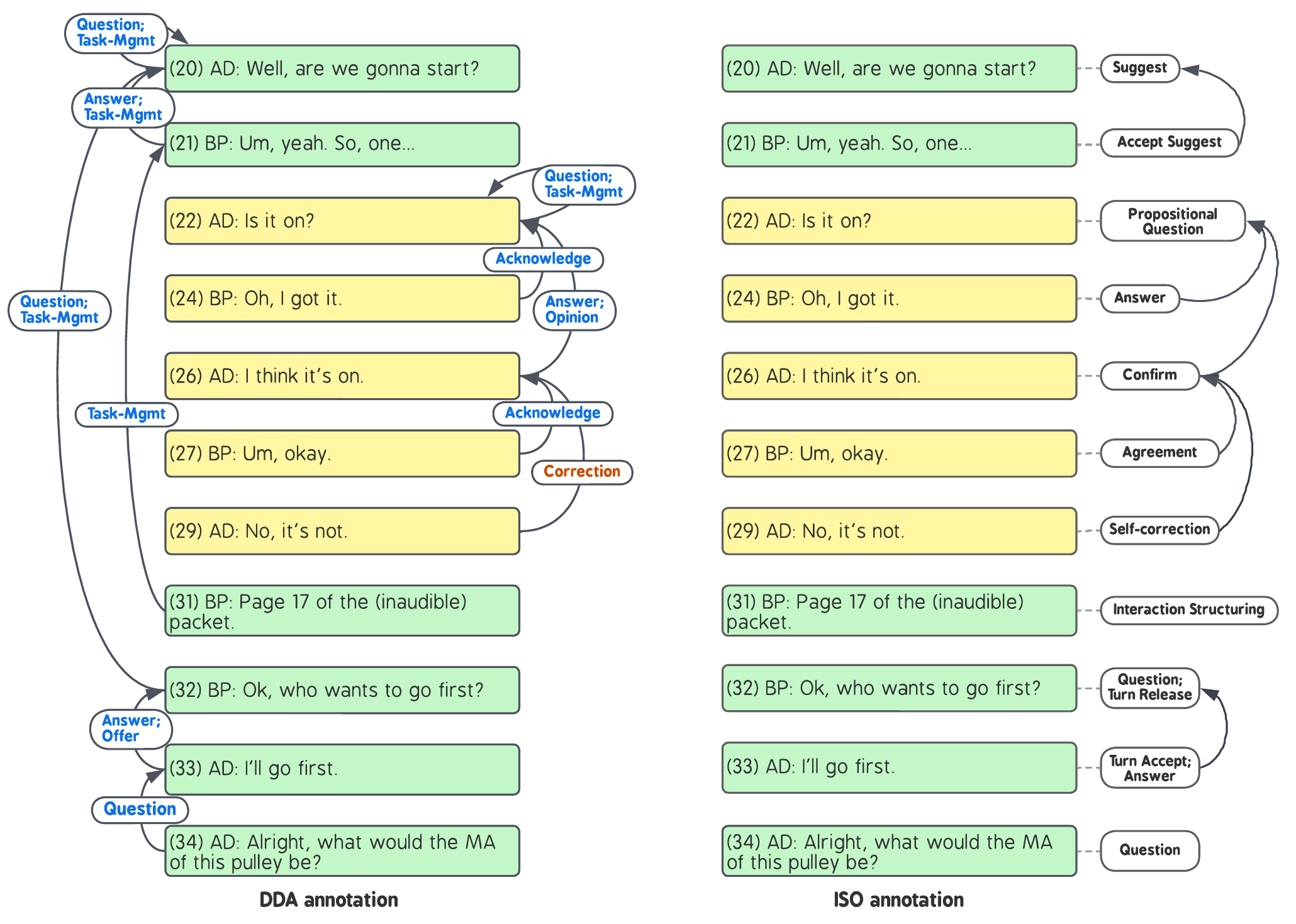}
\caption{A comparison of DDA annotation and corresponding ISO annotation for an example from our classroom setting. Students participate in two interleaved dialog threads: preparing to discuss solutions to a worksheet on pulleys (green) and ensuring their equipment is on correctly (yellow). Response dependencies in DDA connect thread slash units by construction. Functional and similar dependencies in the ISO annotation yield only partial connectivity.\label{fig:dda-iso-classroom}}
\end{figure}
\begin{figure}[t]
    \centering
    \includegraphics[width=\columnwidth]{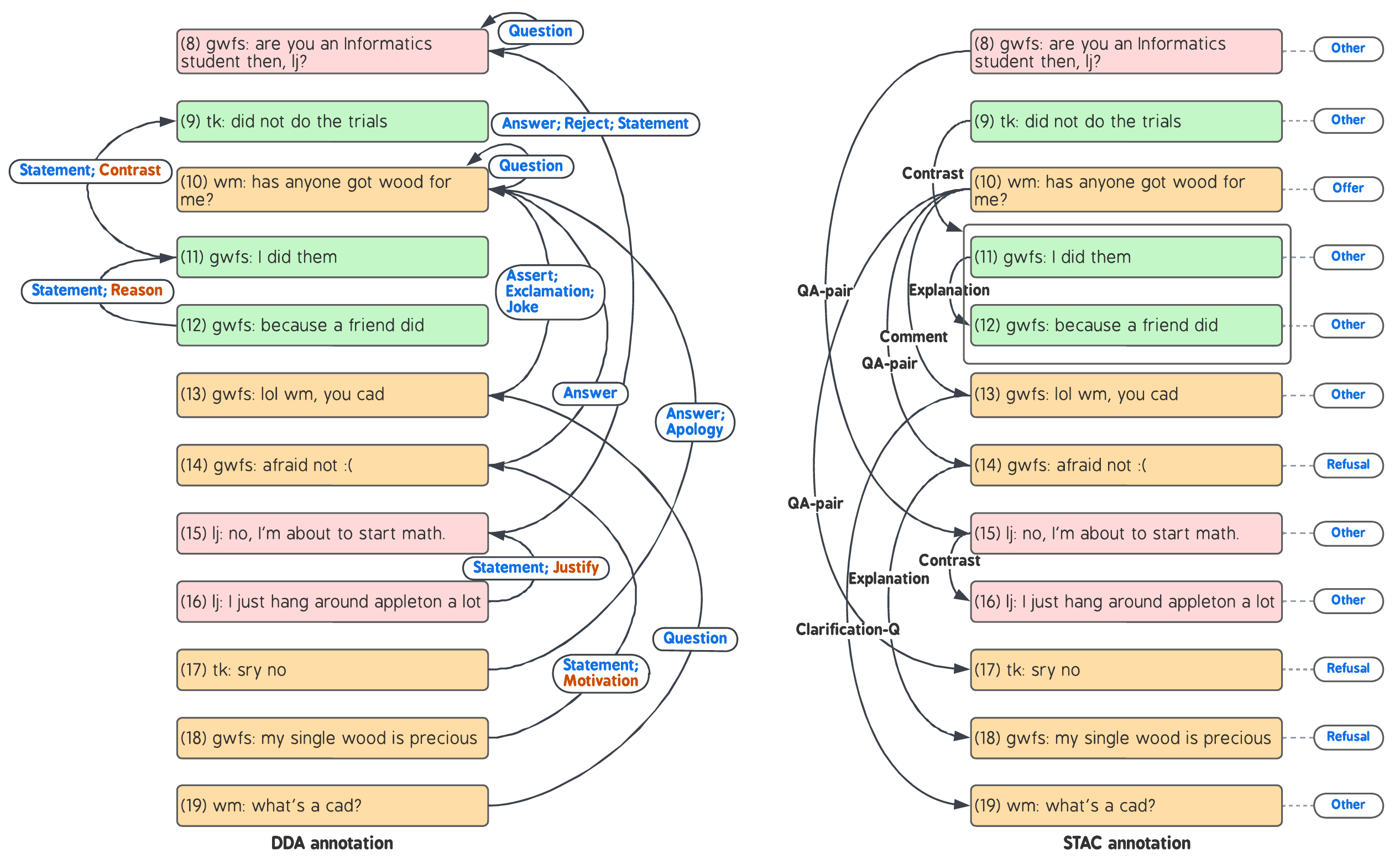}
    \caption{A comparison between DDA annotation (left) and SDRT as adapted for the STAC corpus (right), on an example from the STAC corpus \cite{asher-etal-2016-discourse}. The dialog consists of three distinct threads, denoted by color. Due to the high degree of rhetorical organization in this dialog, both annotations form distinct connected threads. While the relation set in the STAC corpora is tailored for strategic dialog, DDA is able to capture more general phenomena with greater specificity, such as the labeling of apologies within answers or of jokes as opposed to comments.\label{fig:catan-offi}}
    
\end{figure}

\noindent\textbf{Conversation Threads:} Despite the fact that dialogue annotations in DDA might include more structural links than in reply structure graphs, they share the same useful property in which separable conversation threads form connected components in the resulting graph. Given a complete dialogue annotation, this allows a simple method for disentangling threads, a processing step that has been shown to improve dialogue understanding methods \cite{du_discovering_2016} and is of analytical interest in our classroom multi-party dialogue setting. See Fig.~\ref{fig:thread_example} for an example. The dependency chains (8)-(15)-(16), (9)-(11)-(12) and (10)-(13)-(14)-(17)-(18)-(19) can be naturally derived from DDA's response dependency structure. Similarly, in Fig.~\ref{fig:dda-iso-classroom}, we show a classroom example, in which the threads of conversation among students are naturally disentangled by following the response dependencies.\\



\noindent\textbf{Response Dependencies for Discourse Analysis:} The theoretical benefits of DDA's response dependency structure go beyond the threads disentanglement and annotation simplification. DDA can be potentially used as an analytical tool to identify interpersonal relationships and power dynamics. For example, if DDA dependencies show significantly more connections between certain participants, it may indicate they are having more engaged conversations and forming bonding.  Further, if the topological structure of the DDA for a conversation shows balanced connectivity between speakers, it could indicate the power is evenly distributed. Alternatively, if the dependencies are mostly pointing at a single or a few people, it's more likely that they are leading the conversation. We aim to explore these analyses in future work. 
In Table~\ref{table: framework-comparison}, we compare features among different annotation schemes.

\section{Related Work}
\label{sec:related-work}

\noindent\textbf{Dialogue Acts:} There is a long history of analyzing the ``actions" of utterances, known as dialog acts \cite{wittgenstein2010philosophical,austin1975things, searle1969speech,Allen1980AnalyzingII,Kautz1987AFT}.
Dialog act annotation schemes developed include DAMSL (Dialogue
Act Markup in Several Layers) \cite{allen1997draft, core1997coding}, SWBD-DAMSL \cite{jurafsky1997switchboard,godfrey1992switchboard}, DIT (Dynamic Interpretation Theory) \cite{bunt1999dynamic}, and DIT++ schema~\cite{bunt2006dimensions, bunt2009dit++}.
The ISO 24617-2 standard proposed a semantically-based standard for dialogue annotation, and includes both dialogue acts and the relations between discourse units~\cite{bunt-etal-2010-towards, bunt-etal-2012-iso, bunt-etal-2020-iso}.
Researchers have long noted that multi-functionality (pragmatic overloading) is hard to capture with a single utterance purpose, especially in multi-party multi-threaded dialogues~\cite{allwoodactivity,cohen1990intentions,hancher1979classification,di1996pragmatic}.

In our work, we follow SWBD-DAMSL's approach by augmenting its flattened DA tag set. Nevertheless, we made two augmentations: first, DDA handles multi-functionality phenomenon with multi-label and multi-dependency; second, DDA resolves the response structure, which not only unveils a deeper discourse structure in conversations but also anchors the dialogue act and discourse relations into context, which is fundamentally different than tagging schemes. In limited experiments, we find the efficiency of annotating DDA to be comparable to SWBD-DAMSL.\\




\noindent\textbf{Rhetorical Relations in Multi-party Dialogue:} Previous work on structured analysis for multi-party dialogue mainly focused on simple thread disentanglement, rather than analyzing the resulting rhetorical structures~\cite{kummerfeld_large-scale_2019,elsner_you_2008,wang2010making,wang2011learning}. In the current work, we mainly focus on the rhetorical structures in multi-party dialogue.
Four of the most influential frameworks have been used in dialogue analysis: Rhetorical Structure Theory (RST)~\cite{mann1988rhetorical}, Segmented Discourse Representation Theory~(SDRT)~\cite{asher2003logics}, Hobbs' theory of discourse \cite{hobbs1990literature}, and Penn Discourse Treebank (PDTB) framework\cite{prasad-etal-2008-penn,AB2/SUU9CB_2019}. 
In RST, an RST tree is built recursively by connecting the adjacent discourse units, forming a hierarchical structure covering the whole text. RST Bank~\cite{carlson2003building} created a reference corpus for community-wide use, while~\cite{stent-2000-rhetorical} provides a practical analysis on annotating dialogue with RST. Similar to RST, SDRT also provides a hierarchical structure of text organization with full annotation. For example, DISCOR corpus~\cite{reese2007reference}, the ANNODIS corpus~\cite{afantenos2012empirical}, and the STAC~\cite{asher-etal-2016-discourse} use directed acyclic graphs that allow for multiple parents, but not for crossing. 
Based on Hobbs’ theory, Discourse Graphbank~\cite{wolf2005representing} allows for general graphs that allow multiple parents and crossing. Unlike the above frameworks, PDTB adopts a theory-neutral approach to the annotation, which does not aim at achieving complete annotation of the text but focuses on local discourse relations anchored by structural connectives or discourse adverbials. This theory neutrality makes no commitments to what kinds of high-level structures may be created from the low-level annotations of relations and their arguments, thus it permits more freedom of investigating complex dependency structures in multi-party multi-threaded dialogue. Furthermore, ISO DR-Core also follows the theory-neutral stance in PDTB, annotating only high-level, coarse-grained discourse relations that can then be annotated further to capture a finer-grained tree or graph structure, depending on one’s theoretical preferences.

DDA follows PDTB's discourse relation taxonomy since it has been demonstrated effective in annotation practice to yield good annotator agreement, however, we augmented it with dense response structure annotation instead of partial annotation~(as shown in Fig.~\ref{fig:dda-iso}).\\

\noindent\textbf{Annotating Response Structure Graphs:} Another line of work aims to improve conversational understanding systems by uncovering the response dependencies between utterances in multi-party speech \cite{aoki_wheres_2006} or online chat \cite{elsner_you_2008,kummerfeld_large-scale_2019,du_discovering_2016}. Given dialogue segmented into utterances, the task is to connect an utterance of interest with all previous utterances to which it responds. The resulting connected components form dialogue threads that can be understood individually in downstream systems. \cite{kummerfeld_large-scale_2019} present one of the largest available corpora annotated with reply structure graphs, consisting of 77,653 messages from the Ubuntu Internet Relay Chat (IRC). Our notion of response dependency is similar to this line of work with three key differences: (1) All of our dependencies are labeled with the dialogue act and/or rhetorical relation initiated in the responding utterance. (2) All non-self DDA edges point towards previous utterances, but with the duality coding mentioned in Section~\ref{sec:DADR_classes}, the edges still denote the semantic roles of each utterance. (3) An utterance of interest can respond to any number of previous utterances using any number of labels. 

\section{Future Work}

In the future, we plan to apply DDA to annotate multiparty conversations, including conversations from K-12 classrooms, where students form small groups to solve problems collaboratively. \\


\noindent\textbf{Limitations}:
DDA, like all other discourse-level annotation schemes, has its own limitations in terms of scope, generalizability and domain-specific bias. First, DDA assumes sufficient information exists for annotation in the conversation records. If certain references in the context is meant to be resolved from non-verbal communication channels, such as pointing and gestures, DDA may need situated transcripts of the conversation to be properly deployed. Besides, DDA inherits the limitation of the expressive power from PDTB and SWBD-DAMSL and benefits from their scalability in practice. Second, the addressee information is not guaranteed to be reflected in DDA solely. 

\bibliography{custom,dialog-act,discourse,DDA_IWSDS_2023}
\bibliographystyle{spmpsci}

\end{document}